\documentclass[10pt,twocolumn,letterpaper]{article}
\usepackage{iccv}
\usepackage{times}
\usepackage{epsfig}
\usepackage{graphicx}
\usepackage{amsmath}
\usepackage{amssymb}
\usepackage{xcolor}
\usepackage{cuted}
\usepackage{capt-of}
\usepackage{verbatim}
\usepackage{multirow}
\usepackage{makecell}
\usepackage{booktabs}
\usepackage{microtype}
\usepackage{url}
\usepackage{color}
\usepackage{cuted}
\usepackage[accsupp]{axessibility}

\usepackage[pagebackref=true,breaklinks=true,letterpaper=true,colorlinks,bookmarks=false]{hyperref}

\iccvfinalcopy 

\newcommand{\errmetric}{h}
\newcommand{\corr}[1]{\sigma(#1)}
\newcommand{\corrl}[1]{\sigma_l(#1)}
\newcommand{\errR}{\text{Err}_{\mathbf{R}}}
\newcommand{\errt}{\text{Err}_{\mathbf{t}}}
\newcommand{\errpw}{\text{Err}_{\text{pw}}}
\DeclareMathOperator*{\argmin}{arg\,min}

\def\iccvPaperID{8097} 

\ificcvfinal\fi

\begin{document}

\title{A Robust Loss for Point Cloud Registration
}

\author{Zhi Deng\textsuperscript{1}  \qquad \qquad  Yuxin Yao\textsuperscript{1}   \qquad\qquad  Bailin Deng\textsuperscript{2}   \qquad\qquad  Juyong Zhang\textsuperscript{1}\thanks{Corresponding author}\\
\textsuperscript{1}University of Science and Technology of China \quad \qquad \textsuperscript{2}Cardiff University\\
{\tt\small \{zhideng,yaoyuxin\}@mail.ustc.edu.cn  \qquad  DengB3@cardiff.ac.uk \qquad juyong@ustc.edu.cn}
}

\maketitle
\ificcvfinal\thispagestyle{empty}\fi


\begin{abstract}
The performance of surface registration relies heavily on the metric used for the alignment error between the source and target shapes. Traditionally, such a metric is based on the point-to-point or point-to-plane distance from the points on the source surface to their closest points on the target surface, which is susceptible to failure due to instability of the closest-point correspondence. In this paper, we propose a novel metric based on the intersection points between the two shapes and a random straight line, which does not assume a specific correspondence. We verify the effectiveness of this metric by extensive experiments, including its direct optimization for a single registration problem as well as unsupervised learning for a set of registration problems. The results demonstrate that the algorithms utilizing our proposed metric outperforms the state-of-the-art optimization-based and unsupervised learning-based methods.
\end{abstract}

\begin{figure*}[t]
    \begin{center}
       \includegraphics[width=\linewidth]{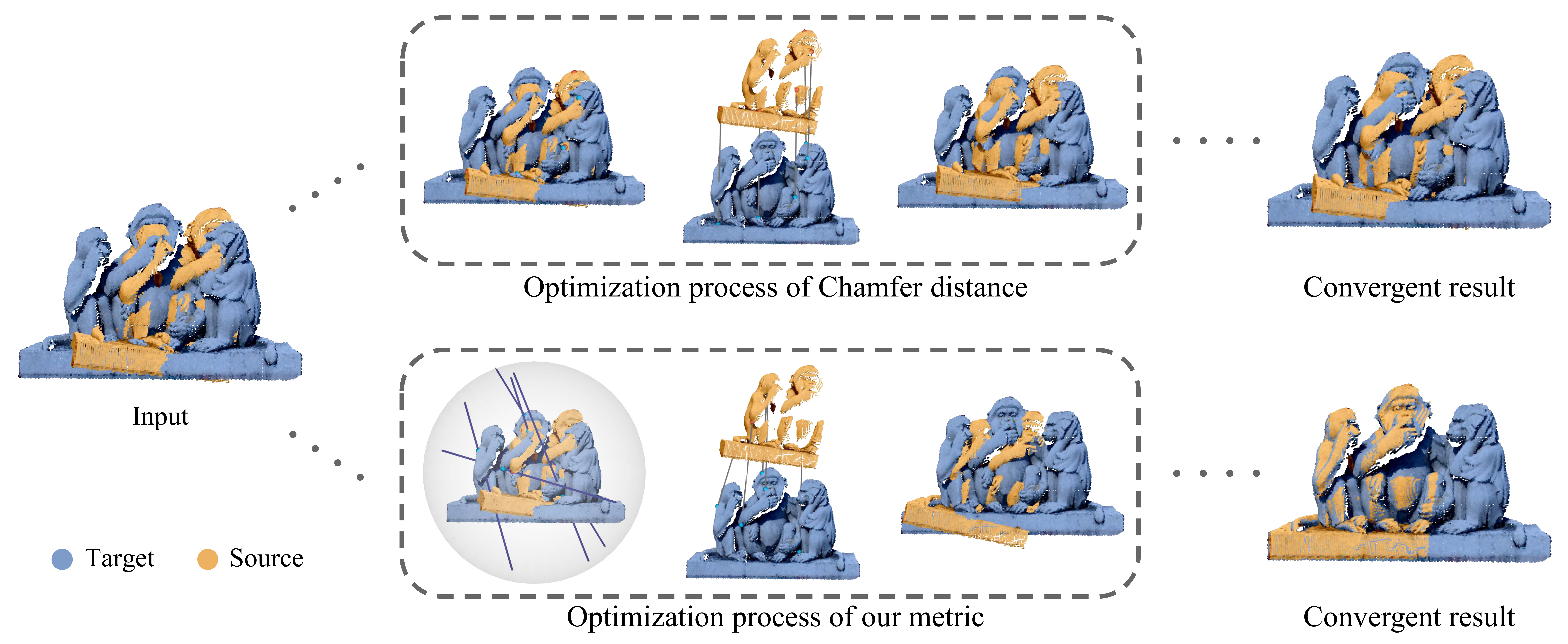}
    \end{center}
    \caption{We propose an error metric for rigid registration based on the intersection between the input shapes and random straight lines that are uniformly distributed. Top: registration by minimizing an error metric based on the Chamfer distance leads to a sub-optimal result. Bottom: with our new metric, the optimization becomes more robust to the local minimum and identifies the correct alignment.}
    \label{fig:introduce our loss}
\end{figure*}


\section{Introduction}
Rigid registration aligns a source shape $\mathcal{S}$ with a target shape $\mathcal{T}$ by applying a rigid transformation $(\mathbf{R}, \mathbf{t})$, where $\mathbf{R}\in\mathbb{R}^{3\times 3}$ is a rotation matrix and $\mathbf{t}\in\mathbb{R}^{3}$ is a translation vector. It is an important task in numerous applications such as 3D scene reconstruction and localization. The transformation is often computed by minimizing a function that measures the alignment error.
In practice, the shapes are often represented as point clouds, and the alignment error is measured using a distance metric $D(\cdot, \cdot)$ evaluated between the points on the source surface and their corresponding points on the target surface:
\begin{equation}
\label{Eq:matching_metric}
\errmetric(\mathbf{R}, \mathbf{t}) = \sum_{(\mathbf{x}, \mathbf{y})\in \mathcal{C}}D(\widetilde{\mathbf{x}},\mathbf{y}),
\end{equation}
where $\mathcal{C}$ is the set of corresponding points between $\mathcal{S}$ and $\mathcal{T}$, and $\widetilde{\mathbf{x}}$ denotes the new position of $\mathbf{x}$ after the transformation. To perform registration in this way, we must first define the corresponding point. Many traditional methods such as the iterative closest point (ICP) algorithm~\cite{Besl_ICP} defines $\mathbf{y}_{\corr{i}}$ as the current closest point to $\mathbf{x}_i$, which needs to be updated in each iteration along with the transformation. It is easy for such iterations to fall into a local optimal solution, especially when noises, outliers, and partial overlaps in the point clouds. Some methods~\cite{Huang_2017_Sys_Structures, Rusinkiewicz_2001_partial, Fitzgibbon_2002_robust_2d_3d, Zhou_2016_FGR} compute local shape descriptors for some sample points, and find the corresponding point on the target surface by matching the descriptors. However, the ambiguity of these hand-crafted descriptors can make them challenging to match,  especially for the point clouds with noises and outliers.

Besides point correspondence, another key component of the alignment error measure in Eq.~\eqref{Eq:matching_metric} is the distance metric $D(\cdot, \cdot)$ between the corresponding pairs. Traditional ICP methods~\cite{Besl_ICP, chen1992object} use the $\ell_2$-norm of the point-to-point or point-to-plane distance as the metric, where $D(\mathbf{x}_i,\mathbf{y}_{\corr{i}})$ is the squared Euclidean distance from $\mathbf{x}_i$ to $\mathbf{y}_{\corr{i}}$ or to the tangent plane of $\mathcal{T}$ at $\mathbf{y}_{\corr{i}}$. To accommodate noise, outliers, and partial overlaps, other methods~\cite{sparseicp_sgp13, bergstrom2014robust, Zhou_2016_FGR, Zhang:2021:FRICP} applied a robust function to the distance values to disregard or down-weight erroneous corresponding pairs. Although such strategies are more robust against noise and partial overlaps, they still rely on the correct point correspondence to some extent. 

In this work, we propose an alignment error metric that does not rely on accurate point correspondence. Our key idea is to intersect the source and target shapes with a random straight line that is uniformly distributed in their bounding sphere. We locate the intersection points from the source shape and the target shape, and use the distance between them as a proxy for the alignment error. We apply Welsch's function~\cite{Paul_Robust_Welsch} to the distance values to obtain a robust measure, and compute its expected value as our alignment error metric. Different from traditional methods, our approach does not assume a specific correspondence rule while still attaining rich information about the alignment from multiple directions thanks to the uniform distribution of the straight lines. Using our metric, optimization-based registration is less susceptible to getting stuck at a local minimum and more likely to obtain a robust solution.

Recently, various deep learning-based approaches for rigid registration have been proposed~\cite{Tejas_2020_Correspondence, Wang_2019_ICCV, Wang_2019_NeurIPS, Huang_2020_CVPR}. However, most of them train the network in a supervised manner and require ground-truth alignment. Our proposed metric can also be used to replace the ground-truth labels and allow a supervised framework to be trained on unlabeled data. It also can fine-tune the model trained by a supervised metric for use on unlabeled datasets in the real world.

In summary, the main contributions of our work are:
\begin{itemize}
	\item We propose a novel error metric for rigid alignment based on intersections between the input shapes and a uniform random straight line, which can improve the robustness of optimization-based rigid registration.
	\item We use the proposed metric to turn various supervised learning frameworks into unsupervised ones that can be trained on real unlabeled data.
\end{itemize}

\section{Related works}
\paragraph{Geometry Processing Using Line Intersection}
Intersection with straight lines has been utilized to process and analyze geometric shapes in the past. In~\cite{LI2003771, LIU200655}, the authors used intersections with random straight lines to compute surface areas of geometric shapes from the perspective of integral geometry~\cite{sankac:2004}. In~\cite{Rovira:2005:PSU}, a method was proposed to sample a point cloud by intersecting with uniformly distributed straight lines. To the best of our knowledge, our work is the first to perform shape registration using intersections with random straight lines.

\paragraph{Optimization-based Registration} 
A classical registration method is Iterative Closest Point (ICP) algorithm~\cite{Besl_ICP}, which obtains the optimal transformation by alternately finding the closest points and updating the transformation. Many variants of ICP have been proposed to improve its efficiency~\cite{chen1992object, Rusinkiewicz_2001_partial, pavlov2018aa,Rusinkiewicz2019}. Another issue of the classical ICP is its robustness to outliers and partial overlaps that often occur in real-world data. Some methods tackled this issue by disregarding some point pairs using heuristics based on their distance or normals~\cite{zhang1994iterative, Rusinkiewicz_2001_partial, CHETVERIKOV2005299}. Another popular approach is to use robust metrics such as the $\ell_p$-norm ($p<1$)~\cite{sparseicp_sgp13} or Welsch's function~\cite{Zhang:2021:FRICP} to measure the alignment error and improve robustness.
Others solved the problem from a statistical perspective and aligned the point clouds via their Gaussian mixture representations~\cite{myronenko2010point, jian2010robust}. 
The above approaches formulate registration as an optimization problem and search for a local minimum using a numerical solver, which require proper initialization. Some other methods formulate a global optimization problem and solve it via either branch-and-bound~\cite{Yang_2016_Go_ICP} or semi-definite relaxation~\cite{maron2016point, dym2017ds++, le2019sdrsac}. They are often computationally more expensive, especially on large-scale problems. Some methods align point clouds by matching their local shape descriptors~\cite{Gelfand_2005_RGR, Rusu_2008_FPH, Rusu_2009_FFPH}. However, the quality of such hand-craft descriptors can be affected by the point density and outliers.

\paragraph{Learning-based Registration} 
Recently, various deep learning approaches have been proposed for registration.
PointNetLK~\cite{yaoki2019pointnetlk} uses an iterative framework which combines the PointNet feature~\cite{qi2017pointnet} and Lucas-Kanade algorithm~\cite{Lucas_1981_LK}. DCP~\cite{Wang_2019_ICCV} utilizes a sub-network to address difficulties in the classical ICP pipeline, which improves the point cloud's features by using DGCNN~\cite{wany_2019_dgcnn} to extract and merge local features. RPM-Net~\cite{yew2020-RPMNet} extracts the hybrid features by learning from spatial coordinates and local geometry, and uses the differentiable Sinkhorn layer and annealing to obtain soft correspondence. PR-Net~\cite{Wang_2019_NeurIPS} uses Gumbel–Softmax with straight-through gradient estimation to obtain a sharp and near-differentiable mapping function. MFG~\cite{wang2020multifeatures} combines the shape features and the spatial coordinates to guide correspondence search independently and fuse the matching results to obtain the full matching. DGR~\cite{choy2020deep} uses a differentiable framework for pairwise registration of real-world 3D scans, adding an optimization module to fine-tune the alignment produced by the weighted Procrustes solver. 
All of the above approaches train their models in a supervised manner, which restricts their applications on real-world unlabeled data.
Recently, FMR~\cite{Huang_2020_CVPR} takes a semi-supervised approach for point cloud registration, by minimizing a feature-metric projection error. In this paper, we propose a new alignment error metric that is suitable for unsupervised learning and achieves better results than the one used in FMR.

\section{Algorithm}
\subsection{Problem Statement}
Point cloud registration is generally posed as an optimization problem. Consider two points clouds 
$\mathbf{X} = \{\mathbf{x}_{i}\}_{i=1}^m$ on the source surface $\mathcal{S}$ and $\mathbf{Y} = \{\mathbf{y}_{j}\}_{j=1}^n$ on the target surface $\mathcal{T}$, where $\mathbf{x}_{i}, \mathbf{y}_{j}\in \mathbb{R}^{3}$ are the points. Let $\widetilde{\mathbf{X}}=\{\widetilde{\mathbf{x}}_i\}_{i=1}^m$ denote the deformed source point cloud with the rigid transformation $(\mathbf{R}, \mathbf{t})$, where
\[
\widetilde{\mathbf{x}}_i=\mathbf{R}\mathbf{x}_i+\mathbf{t}.
\]
Using the alignment error given in Eq.~\eqref{Eq:matching_metric}, ICP-based methods can be described as 
\[
(\mathbf{R}^{*},\mathbf{t}^{*})=\mathop{\arg\min}\limits_{(\mathbf{R},\mathbf{t})}\sum_{\mathbf{x}_i\in\mathbf{X}}D(\widetilde{\mathbf{x}}_i, \mathbf{y}_{\corr{i}}),
\]
where $\corr{i}$ denotes the index of the corresponding point in $\mathbf{Y}$ for the point $\mathbf{x}_i \in \mathbf{X}$.
The above formulation only considers the distance from the source point cloud to the target point cloud. 
Considering the distance from the target to the source as well, the Chamfer distance has also been used to measure the deviation between two point clouds~\cite{Fan2017, Huang_2020_CVPR}.
The alignment error based on the Chamfer distance can be written as: 
\begin{equation}
\label{Eq:chamfer_distance}
\errmetric(\widetilde{\mathbf{X}}, \mathbf{Y})=\sum_{\mathbf{x}_i\in\mathbf{X}}D(\widetilde{\mathbf{x}}_i, \mathbf{y}_{\corr{i}}) + \sum_{\mathbf{y}_j\in\mathbf{Y}}D(\widetilde{\mathbf{x}}_{\rho(j)}, \mathbf{y}_{j}),
\end{equation} 
where ${\rho{(j)}}$ denotes the index of the corresponding point in $\mathbf{X}$ for the point $\mathbf{y}_j \in \mathbf{Y}$. The choices of ${\sigma{(\cdot)}}$ and ${\rho{(\cdot)}}$ can affect the quality of registration. In ICP-based methods, $\mathbf{y}_{\corr{i}}$ is chosen to be the closest point to $\mathbf{x}_i$. But such closest-point correspondence is often incorrect when there is large misalignment or a low overlap ratio between the two point clouds. 
Therefore, we would like to use an alignment error that does not presume a pre-defined rule of point correspondence while still being effective in guiding the alignment. Our key observation is that for two shapes that are perfectly aligned, any straight line that intersects with one shape will also intersect the other shape at the same points. When the two shapes are close, their intersection points with the same line will also be close to each other. Moreover, if we use a set of random straight lines to intersect with the two shapes, then the intersection points along each line can inform us about the difference between the two shapes from a particular viewpoint along a view ray that corresponds to the line. In the past, such random straight lines have been utilized in integral geometry to determine geometric properties of a given shape such as surface area~\cite{LI2003771, LIU200655}. In the following, we propose an alignment error metric based on the intersection with random straight lines.

\subsection{Error Metric Based on Line Intersection}
To measure the alignment error between a source shape $\mathcal{S}$ and a target shape $\mathcal{T}$, our basic idea is to intersect both shapes with a set of random straight lines with a uniform distribution, and compare the intersection points along each line. Specifically, given a straight line $l$ that intersects with both shapes, we denote the set of intersection points with the source shape and the target shape as  $\mathcal{S}_l=\{\mathbf{x}^l_i\}$ and $\mathcal{T}_l=\{\mathbf{y}^l_j\}$, respectively. Then we measure the deviation between the two sets of intersection points as: 
\begin{equation}
\label{Eq:line_metric}
F_l(\mathcal{S}, \mathcal{T})= w_l \left(\sum_{\mathbf{x}_i^l\in\mathcal{S}_l}
D({\mathbf{x}}_i^l, \mathbf{y}_{\corrl{i}}^l) 
+\sum_{\mathbf{y}_j^l\in\mathcal{T}_l} D({\mathbf{x}}_{{\rho_l(j)}}^l,\mathbf{y}_j^l)\right),
\end{equation}
where $D$ is an error metric that will be explained later, and 
$$\corrl{i} = \argmin\limits_{k} \| \mathbf{x}_i^l - \mathbf{y}_k^l \|, ~~ \rho_l(j) = \argmin\limits_{k} \| \mathbf{x}_k^l - \mathbf{y}_j^l \|, $$
i.e., $\mathbf{y}_{\corrl{i}}^l$ is the closest point in $\mathcal{T}_l$ to $\mathbf{x}_i^l$, and $\mathbf{x}_{\rho_l(j)}^l$ is the closest point in $\mathcal{S}_l$ to $\mathbf{y}_j^l$. The weight $w_l$ is defined as $w_l = \exp(- |\frac{|\mathcal{S}_l| - |\mathcal{T}_l|}{2}|)$. This reduces the weight for a line with a large difference between the numbers of its intersection points with the two shapes, which may indicate an erroneous correspondence between them along the line. Finally, the alignment error between $\mathcal{S}$ and  $\mathcal{T}$ is defined as the expected value of $F_l(\mathcal{S}, \mathcal{T})$ over the distribution of the lines:
\begin{equation}
    \errmetric(\mathcal{S}, \mathcal{T}) = E(F_l(\mathcal{S}, \mathcal{T})).
    \label{Eq:our_metric}
\end{equation}
To apply this in point cloud registration, we evaluate the error metric in Eq.~\eqref{Eq:our_metric} for the transformed source point cloud $\widetilde{\mathbf{X}}$ and the target point cloud $\mathbf{Y}$, and use it as the target function for an optimization-based method or as a loss function term for a learning-based approach. In the following, we present the details for evaluating the error metric on point clouds.

\paragraph{Choice of $D$ in Eq.~\eqref{Eq:line_metric}}
With Eq.~\eqref{Eq:line_metric}, we effectively establish correspondence between points on the source and target shapes along a straight line. However, since the line is chosen randomly, the correspondence may be inaccurate. Therefore, we define $D$ to be a robust metric to alleviate the impact of inaccurate correspondence as well as outliers. We choose Welsch's function as the metric:
\begin{equation}
D(\mathbf{x}, \mathbf{y}) =  \psi_{\nu}(\|\mathbf{x}-\mathbf{y}\|_{2}),
\label{Eq:sparse_metric}
\end{equation}
where $\psi_{\nu}(x)=1-\exp(-\frac{x^2}{2{\nu}^{2}})$, and $\nu > 0$ is a parameter. To take into account the scale of input point clouds, we set $\nu =\nu_{0} d_{\text{med}}$, where ${d}_{\text{med}}$ is the median distance between all corresponding point pairs and $\nu_{0}$ is a user-specified parameter. We choose $\nu_{0}$ in all experiments.
We treat $\nu$ as a constant term during optimization/training and do not evaluate/back-propagate its gradient. The value $d_{\text{med}}$ is updated in each iteration according to the latest alignment.

\paragraph{Generation of Random Straight Lines}
Following~\cite{LIU200655}, we first compute a bounding sphere $S_{r}$ that covers both the source and the target point clouds. Then we sample two independent uniformly distributed points on $S_{r}$ and connect them to generate a random straight line. Each point on the sphere can be parameterized as: 
\[
S_r(u, \alpha)=(r\sqrt{1-u^2}\cos{\alpha}, r\sqrt{1-u^2}\sin{\alpha}, ru),
\]
where $r$ is the radius, $u\in[-1,1]$, and $\alpha\in[0, 2\pi)$. The random points on the sphere are generated by uniformly sampling the parameters $u$ and $\alpha$ in their domains~\cite{LIU200655}.
In each iteration of optimization or training, we use this approach to generate 15000 straight lines.

\begin{figure}[t]
	\begin{center}
		\includegraphics[width=\linewidth]{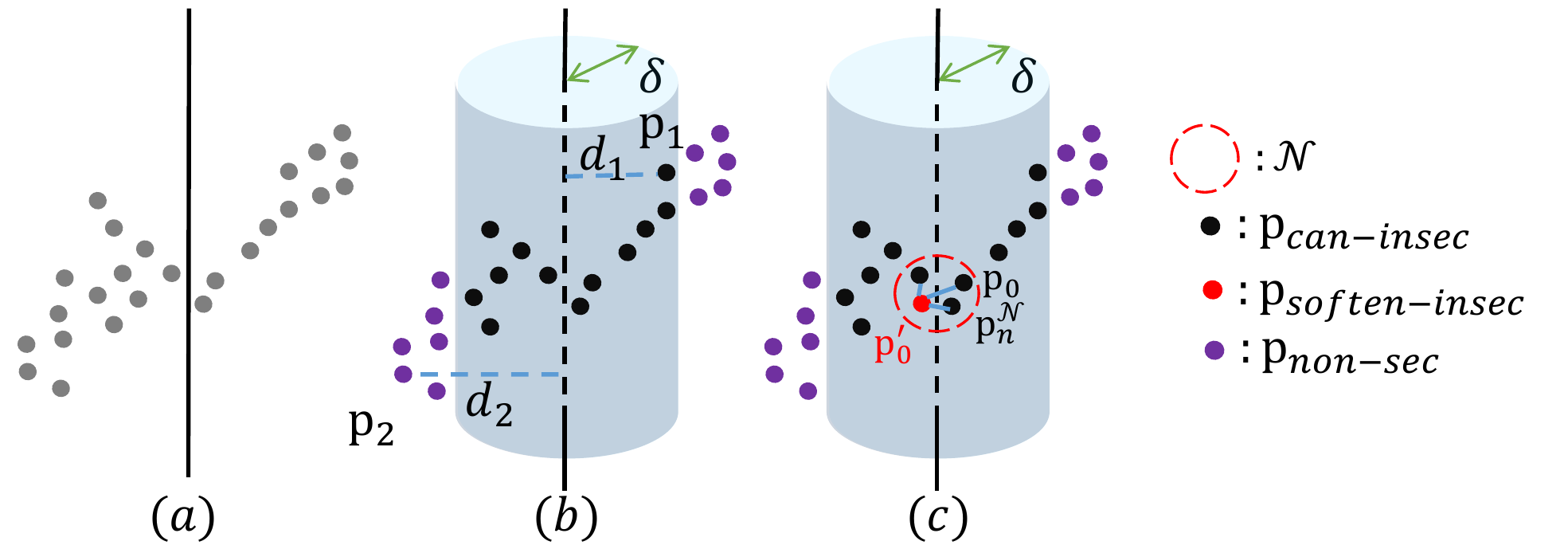}
	\end{center}
	\caption{Illustration of the process of generating the intersections between the point cloud and a straight line.  
		(a) shows the local point cloud and the straight line; (b) shows the selection of the candidate points (black dots); (c) shows the process of softening the candidate points to obtain the final intersections (red dot).}
	\label{fig:distributed the intersected lines}
\end{figure}

\paragraph{Line Intersection with Point Clouds}
Since a point cloud contains discrete samples of the underlying shape, a straight line that intersects with the underlying shape does not necessarily intersect with the points in the point cloud. Therefore, we use the following steps to approximate the intersection between the straight line and the underlying shape (see Fig.~\ref{fig:distributed the intersected lines}). First, similar to~\cite{LIU200655}, we enlarge the line into a cylinder that is centered at the line and has radius $\delta$, and include all the points contained within the cylinder as candidate points for the intersection (Fig.~\ref{fig:distributed the intersected lines}~(b)). Then for each candidate point $\mathbf{p}_{0}$ whose $k$-nearest neighbors in the point cloud are also candidate points, we compute a convex combination of $\mathbf{p}_{0}$ and it's $k$-nearest neighbors as an intersection point (Fig.~\ref{fig:distributed the intersected lines}~(c)):
\begin{equation}
\mathbf{p}_0'=\frac{\sum_{\mathbf{p}\in \mathcal{N}(\mathbf{p}_{0})}d_{\mathbf{p}}\mathbf{p}}{\sum_{\mathbf{p}\in \mathcal{N}(\mathbf{p}_{0})}{d_{\mathbf{p}}}},
\label{eq:IntersectionCombination}
\end{equation}
where the set $\mathcal{N}(\mathbf{p}_{0})$ contains $\mathbf{p}_{0}$ and its $k$-nearest neighbors, and  $d_{\mathbf{p}}$ is the distance from the point $\mathbf{p}$ to the line.
In our implementation, we set $k=2$, and choose $\delta = \frac{\sqrt{3}}{2}d_{\text{nei}}$ where $d_{\text{nei}}$ is the average distance across the whole point cloud between a point and its $k$-nearest neighbors.

\section{Experiments}
In this section, we apply our alignment error metric to different registration problems, include both optimization-based and learning-based methods, and test it on on both synthetic and real datasets to verify its effectiveness.

\subsection{Datasets} \label{section.4.1}

\paragraph{Synthetic Datasets} Our synthetic datasets are generated from the ModelNet40 dataset~\cite{Wu_2015_CVPR} and the Axyz-pose human dataset~\cite{Axyz}. The ModelNet40 dataset contains CAD models from 40 artificial object categories. We select 625 cases from the Airplane category to construct an Airline dataset, randomly sampling 500 cases for training and using the remaining 125 for testing. The Axyz-pose human dataset contains 110 clothed human mesh models. We randomly choose 110 models to construct a Human dataset, using 100 models for training set and the remaining 10 for testing.

Using the datasets above, we generate point cloud pairs for training and testing. To make the generated point clouds similar to the type of data captured by an RGBD camera, we use the following steps to generate the data.
Firstly, we sample a complete model from different perspectives to generate partially overlapping data. Specifically, we choose a certain axis and rotate an imaginary camera around it to derive $N$ camera locations with rotation angles at regular intervals (we set $N$ to $50$ for the Human dataset and $18$ for the Airplane dataset). We then save the visible part of the model from each camera location as a source point cloud. For each source point cloud of the Airplane dataset, we rotate the camera by a random angle in the range $[-75^{\circ}, 75^{\circ}]$ with respective to a random axis, and save the visible part as the corresponding target point cloud. For the Human dataset, we use the whole model to construct the target point cloud. Then we scale each point cloud pair to be contained within $[-1, 1]^{3}$. 
Secondly, we generate composite transformations between the source and target point clouds. We follow~\cite{Wang_2019_ICCV} to generate rotations by sampling three Euler angle rotations in the range $[0, 45^{\circ}]$ and translations in the range $[-0.2, 0.2]$ on each axis. In total, the Airplane dataset contains 9000 pairs for training and 2250 for testing, while the Human dataset contains 5000 pairs for training and 500 for testing.
For each point cloud, we use PCL to compute the point normals~\cite{PCL}, and use FPS~\cite{Point++} to sample 1024 points.

 \paragraph{Real Dataset} To test our metric on unlabeled data, we also construct a real dataset based on the 3D-Match dataset~\cite{zeng20163dmatch}, the 7scenes dataset~\cite{shotton2013scene} and the RGB-D SLAM dataset~\cite{sturm12iros}. Inevitably, our metric cannot handle point cloud pairs with arbitrary pose differences and overlap ratios. And in practice, it is uncommon to have extremely large differences in poses or extremely small overlap ratios. Therefore, we select point cloud pairs separated by 20 frames from the RGB-D SLAM dataset and the 7scenes dataset, respectively. For the 3D-Match dataset, we collect the pre-processed data pairs from~\cite{choy2020deep}\footnote{\url{https://github.com/chrischoy/DeepGlobalRegistration}} where the overlap ratio is greater than $70\%$. All point cloud pairs are scaled into $[-5, 5]^{3}$. Finally, the real dataset is divided into 8000 pairs for training and 2000 pairs for testing. 
 Similar to the synthetic dataset, we compute the point normals and sample 2048 points for each point cloud.

 \begin{table}[t]
 	\caption{Comparison between different optimization-based methods on the Human dataset~\cite{Axyz}.
 	}
 \begin{center}
\setlength{\tabcolsep}{1.8pt}
{\begin{tabular}{cccc}
\toprule  
Method& \makecell[c]{$\errR{}$\\ (degrees)}& \makecell[c]{$\errt{}$ ($\cdot 10^{-1}$)\\ ($\ell_1$, $\ell_2$)} & \makecell{$\errpw{}$ ($\cdot 10^{-1}$)\\($\ell_1$, $\ell_2$)}\\
\midrule  
ICP~\cite{Besl_ICP} & 10.015& 0.139, 0.093&0.112, 0.082\\
FRICP~\cite{Zhang:2021:FRICP} &6.001 &0.096, 0.064&0.074, 0.054 \\
FGR~\cite{Zhou_2016_FGR} &46.31&0.411, 0.274&0.616, 0.41\\
\midrule  
CD&  5.863 &0.148, 0.132&0.151, 0.14\\
CD-W ($\nu_{0} = 0.5$)& {4.84}& { 0.086, 0.078}&{0.108, 0.087}\\
Ours& \textbf{0.576}& \textbf{0.017, 0.013}& \textbf{0.018, 0.015}\\
\bottomrule 
\end{tabular}}
\end{center}
\label{tab:Human dataset optimized}
\end{table}

 \paragraph{Evaluation Criteria} 
 We evaluate the registration accuracy on a point cloud pair using the isotropic rotation error $\errR{}$ and translation error $\errt{}$ inspired by~\cite{yew2020-RPMNet}, as well as the pointwise error $\errpw{}$:
\begin{equation}
\begin{aligned}
&\errR{} =\angle(\mathbf{R}_{\text{GT}}^{-1}\mathbf{\hat{R}}),\quad 
\errt{}=\|\mathbf{t}_{\text{GT}}-\mathbf{\hat{t}}\|_{*}, \\
&\errpw{} = \frac{1}{|\mathbf{X}|}\sum_{\mathbf{x}_i\in\mathbf{X}}\|\mathbf{R}_{\text{GT}}\mathbf{x}_i+\mathbf{t}_{\text{GT}}-\hat{\mathbf{R}}\mathbf{x}_i-\hat{\mathbf{t}}\|_{*},
\end{aligned}
\label{Eq:eval_metric}
\end{equation}
where $\mathbf{R}_{\text{GT}}$ and $\mathbf{t}_{\text{GT}}$ are the ground-truth rotation and translation respectively, $\hat{\mathbf{R}}$ and $\hat{\mathbf{t}}$ are the computed rotation and translation respectively, $\angle(\mathbf{A}) = \arccos(\frac{tr(\mathbf{A}) - 1}{2})$ is the angle of the rotation matrix $\mathbf{A}$ in degrees, $|\mathbf{X}|$ is the number of points in the source point cloud $\mathbf{X}$, and $\|\cdot\|_{*}$ is either the $\ell_1$-norm or the $\ell_2$-norm.
We use the mean values of these metrics to measure the performance of a method on a benchmark dataset.
For the figures in this section and the supplementary material, the number under each result is the pointwise error with the $\ell_2$-norm.

\begin{figure}[t]
    \begin{center}
       \includegraphics[width=\linewidth]{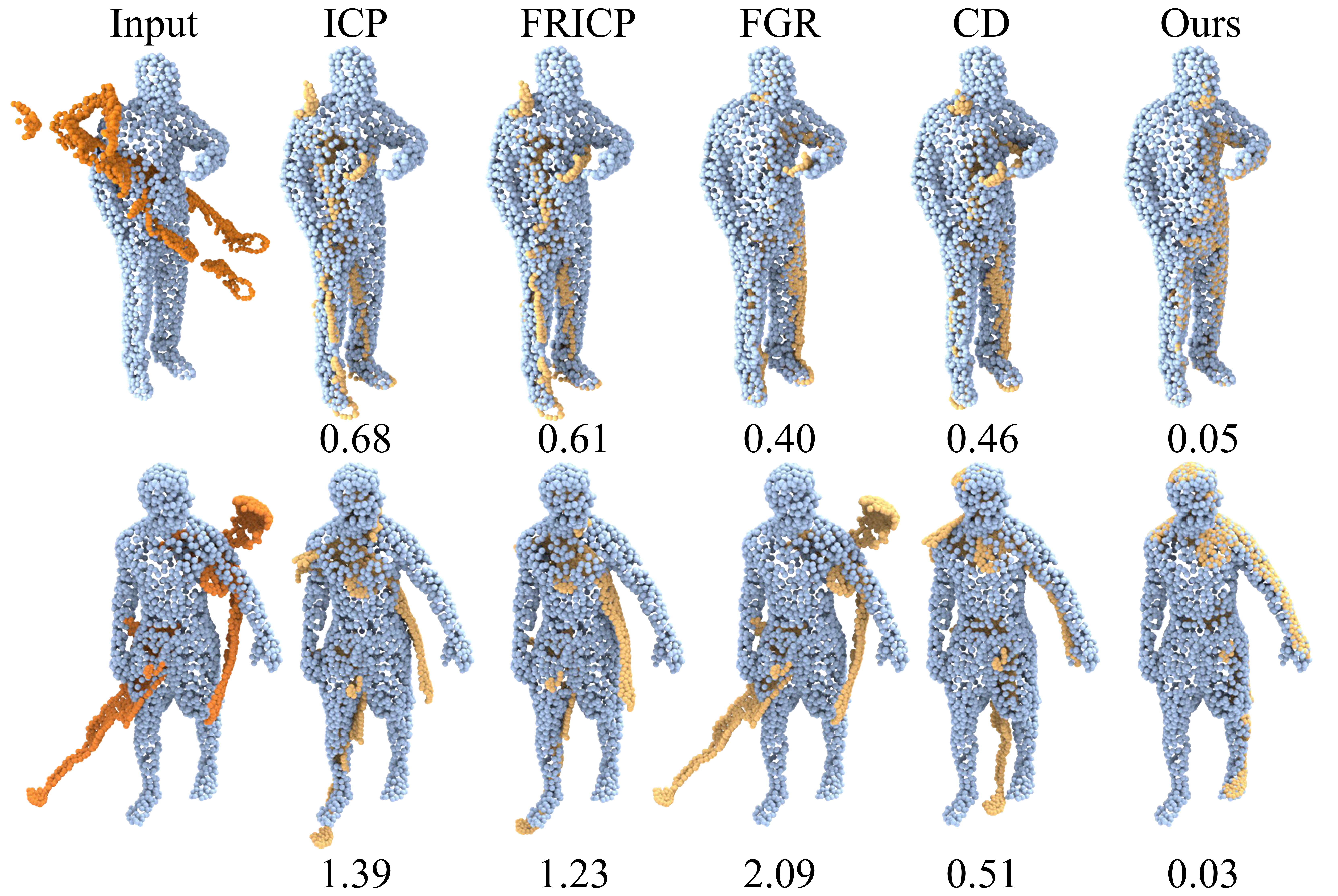}
    \end{center}
    \caption{Comparison of registration results on the Human dataset~\cite{Axyz} using different optimization-based methods. 
    }
     \label{fig: human_optimize results}
 \end{figure}

\subsection{Effectiveness of Our Metric}\label{ablation sec}

\begin{figure}[!t]
	\centering
	\includegraphics[width=\linewidth ]{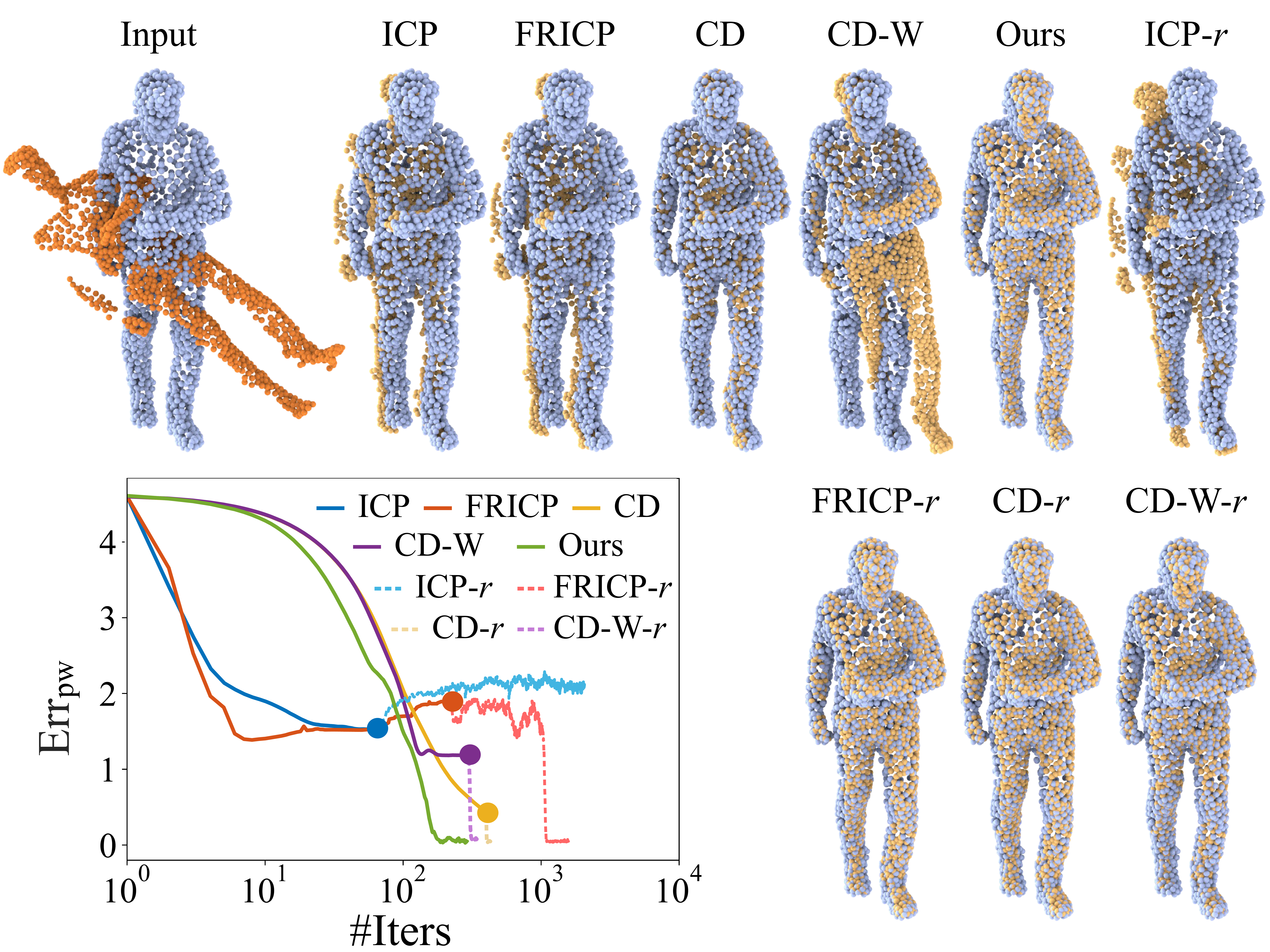}
	\caption{We take convergent results of different optimization-based methods (shown as solid dots in the $\ell_2$ $\errpw{}$-plot) to initialize a minimization of our metric (denoted as $*$-$r$ where $*$ is the original method). In three out of four cases, our optimization escapes from the local minima of the original method and further reduces $\errpw{}$. 
    }
	\label{fig:error_single_case}
\end{figure}

\begin{figure}[!t]
	\centering
	\includegraphics[width=\linewidth ]{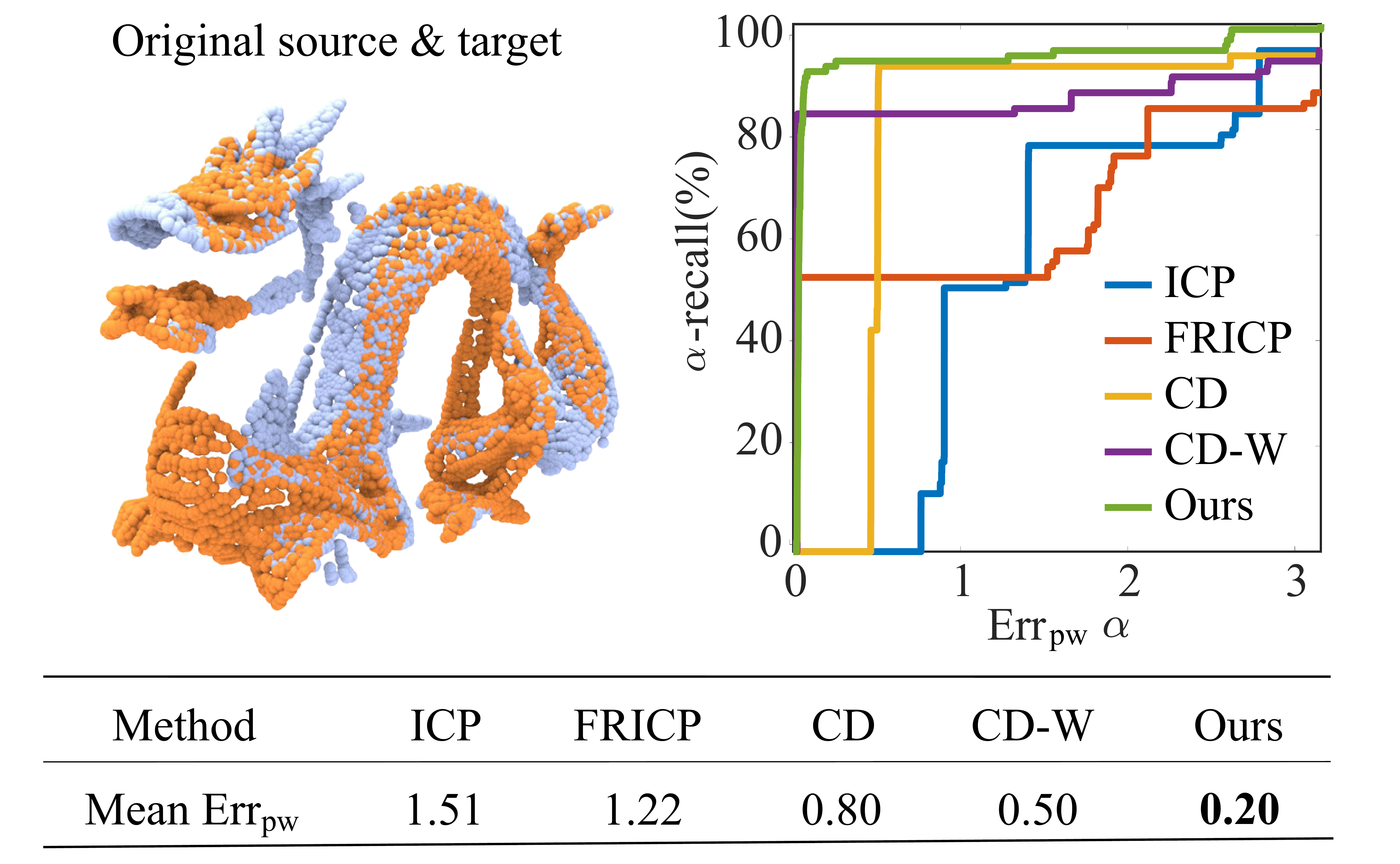}
	\caption{$\alpha$-recall rates and mean $\ell_2$-$\errpw{}$ from different optimization-based methods on a pair of point clouds with 100 random initial alignments. Optimization using our metric is less sensitive to initialization.
    }
	\label{fig:error_single_case2}
\end{figure}

\paragraph{Comparison with Optimization-based Methods}
We optimize the Lie algebraic representation of rigid transformation with our metric as the target function using the Adam optimizer~\cite{kingma2017adam} in Pytorch~\cite{NEURIPS2019_9015}. Using the Human test dataset as the benchmark, we compare our results with other optimization-based methods, including ICP~\cite{Besl_ICP}, FRICP~\cite{Zhang:2021:FRICP} and FGR~\cite{Zhou_2016_FGR} with their open-source implementations {\footnote{\url{https://github.com/yaoyx689/Fast-Robust-ICP}}\footnote{\url{https://github.com/intel-isl/FastGlobalRegistration}}}.
We also compare with two optimization approaches that use the Chamfer distance in Eq.~\eqref{Eq:chamfer_distance} as the target function, with the metric $D$ chosen to be the Euclidean distance (denoted as CD) and Welsch's function in Eq.~\eqref{Eq:sparse_metric} (denoted as CD-W, see also the supplementary material for more details), respectively.
Tab.~\ref{tab:Human dataset optimized} shows the performance of different methods on the Human test dataset. We can see our proposed metric can generate more accurate results than other traditional optimization methods. Fig.~\ref{fig: human_optimize results} shows some examples of registration results from different methods, where other methods converge to sub-optimal a local minimum while our metric leads to more accurate alignment.
Fig.~\ref{fig:error_single_case} and Fig.~\ref{fig:error_single_case2} further illustrate the effectiveness of our metric in avoiding local minimum.
In Fig.~\ref{fig:error_single_case}, we take the convergent results of other optimization-based methods as initialization for optimization using our metric as the target function. The plot of pointwise $\ell_2$ error shows that our approach can often further reduce the alignment error, which indicates its capability to escape from a local minimum of other methods. Fig.~\ref{fig:error_single_case2} demonstrates the robustness of our approach to the initial alignment. We use a point cloud pair generated from the Dragon model in the Stanford 3D Scanning Repository\footnote{\url{http://graphics.stanford.edu/data/3Dscanrep/}}, and apply a random transformation to one of them to derive 100 different initial alignments. Then we perform registration using different methods and compare their $\alpha$-recall rates $|S_\alpha|/|S|$ for different $\alpha$ values, where $|S|$ is the total number of test cases and $|S_\alpha|$ is the number of test cases where the pointwise $\ell_2$ error is
less than $\alpha$~\cite{Zhou_2016_FGR} (i.e., for a given $\alpha$, a larger $\alpha$-recall rate indicates better performance). The $\alpha$-recall plot as well as the mean $\ell_2$ pointwise error show that our method is less sensitive to initialization than other methods.

\label{sec:empirical_evidence}

\paragraph{Ablation Studies}
In Tab.~\ref{tab:Ablation study}, we use the Human test dataset as the benchmark to verify the effectiveness of different components of our metric. The first row shows that the use of convex combination for computing the intersection point in Eq.~\eqref{eq:IntersectionCombination} is important. Here Insec1 denotes an alternative approach where we simply take all the candidate points as the final intersection points, which leads to worse performance than our approach (shown in the last row). The second and the third rows show two alternative sampling approaches (Sample1 and Sample2) for the random straight lines. For Sample1, we uniformly sample a point in the bounding box and uniformly sample a direction, and construct a line that goes through the sampled point along the sampled direction. For Sample2, we sample a point uniformly from the source and the target point cloud, respectively, and make a uniformly small perturbation, and connect them to obtain a straight line. Both approaches are outperformed by our sampling method (the last row). The fourth to the seventh rows show the impact of different $\nu$ values on our metric. For Welsch's function, a  larger $\nu$ makes it closer to the $\ell_2$-norm, whereas a smaller $\nu$ makes it closer to the $\ell_0$-norm. The results show that a choice of $\nu_{0}$ close to $0.5$ is suitable.

  \begin{table}[t]
	\caption{Optimization using our metric with different settings on the Human dataset~\cite{Axyz}, including an alternative line intersection method (Insec1), two alternative line sample methods (Sample1 and Sample2), and different values of $\nu_0$.}
	\begin{center}
		\setlength{\tabcolsep}{3pt}
		{\begin{tabular}{ccccc}
				\toprule  
Method& \makecell[c]{$\errR{}$\\ (degrees)}& \makecell[c]{$\errt{}$ ($\cdot 10^{-1}$)\\ ($\ell_1$, $\ell_2$)} & \makecell{$\errpw{}$ ($\cdot 10^{-1}$)\\($\ell_1$, $\ell_2$)}\\
				\midrule  
				Insec1 ($\nu_{0}=0.5$)& {49.27}& {0.831, 0.744}& {1.175, 0.772}\\
				\midrule  
				Sample1&  20.482 &0.274, 0.175 &0.419, 0.276\\
				Sample2& 5.201& {0.151, 0.093}& {0.117, 0.075}\\
				\midrule  
				$\nu_{0} = 100$& {7.786}& {0.289, 0.191}& {0.292, 0.195}\\
				$\nu_{0} = 10$& {8.181}& { 0.288, 0.190}&{0.291, 0.194}\\
				$\nu_{0} = 1$& {2.814}& {0.044, 0.029}& {0.051, 0.034}\\
				$\nu_{0} = 0.01$& {10.814}& {0.326, 0.217}& {0.334, 0.223}\\
				\midrule
				Ours ($\nu_{0}=0.5$)& \textbf{0.576}& \textbf{0.017, 0.011}& \textbf{0.018, 0.012}\\
				\bottomrule 
			\end{tabular}}
		\end{center}
		\label{tab:Ablation study}
	\end{table}

\subsection{Results for Unsupervised Learning}
We also use the proposed metric for deep learning-based registration. Specifically, we replace the alignment term in the loss in the frameworks of DCP~\cite{Wang_2019_ICCV}, FMR~\cite{Huang_2020_CVPR}, and RPM-Net~\cite{yew2020-RPMNet}, and train them in an unsupervised manner. We compare the results with the original frameworks trained on the same data with ground-truth alignment labels (denoted as DCP-GT, FMR-GT and RPM-GT respectively). For comparison, we also include unsupervised variants of DCP and RPM-Net using the $\ell_2$ Chamfer distance as the alignment term, as well as the semi-supervised version of FMR from~\cite{Huang_2020_CVPR} (denoted as DCP-CD, RPM-CD and FMR-CD respectively). For all frameworks, we replace the batch normalization with the group normalization for better-unsupervised training. For the unsupervised variants of RPM-Net, we set the weights of the regularization term and the alignment term to $10$ and $1$ respectively, and the learning rate to $2\times 10^{-6}$. For the unsupervised variants of DCP, we set the weights of the cycle term and the alignment term to $0.01$ and $1.0$ respectively, and the learning rate to $10^{-5}$. For the unsupervised variant of FMR, we set the weights of the encoder term and the alignment term to $0.001$ and $1$ respectively, and the learning rate to $10^{-5}$. All frameworks are trained using the Adam optimizer from Pytorch for 50 epochs, on a workstation with two Intel Xeon Silver 4110 CPUs at 2.10 GHz, and four Tesla V100 GPUs.

\begin{table}[t]
	\caption{Comparison between different optimization-based and learning-based methods on the Airplane dataset~\cite{Wu_2015_CVPR}.}
	\begin{center}
		\resizebox{0.46\textwidth}{!}{\begin{tabular}{cccc}
				\toprule
Method& \makecell[c]{$\errR{}$\\ (degrees)}& \makecell[c]{$\errt{}$ ($\cdot 10^{-1}$)\\ ($\ell_1$, $\ell_2$)} & \makecell{$\errpw{}$ ($\cdot 10^{-1}$)\\($\ell_1$, $\ell_2$)}\\
				\midrule 
				ICP~\cite{Besl_ICP} & 7.223&0.131, 0.087 &0.136, 0.105 \\
				FRICP~\cite{Zhang:2021:FRICP} & 6.91&0.076, 0.051&0.123, 0.094 \\
				FGR~\cite{Zhou_2016_FGR} &13.72&0.099, 0.065& 0.156, 0.114\\
				\midrule
				DCP-GT& 2.281& 0.067, 0.044& 0.073, 0.05\\
				DCP-CD& 6.612 &0.165, 0.11 &0.198, 0.139\\
				DCP-Ours& \textbf{3.808}& \textbf{0.082, 0.056}& \textbf{0.093}, \textbf{0.063}\\
				\midrule
				FMR-GT& 1.977 & 0.086, 0.055& 0.099, 0.065 \\
				FMR-CD& 5.819 & 0.215, 0.153& 0.247, 0.19\\
				FMR-Ours& \textbf{2.51}& \textbf{0.117, 0.076}& \textbf{0.134, 0.091}\\
				\midrule
				RPM-GT& 2.19& 0.041, 0.034 & 0.043,0.03\\
				RPM-CD& 2.791 &0.103, 0.089 &0.102,0.083 \\
				RPM-Ours& \textbf{1.673}& \textbf{0.042, 0.034} &\textbf{0.045, 0.031}\\
				\bottomrule
			\end{tabular}}
		\end{center}
		\label{tab:Airplane dataset}
	\end{table}

\begin{table}[t]
	\caption{Comparison between different learning-based methods on the Human dataset~\cite{Axyz}.}
\begin{center}
\resizebox{0.46 \textwidth}{!}{\begin{tabular}{cccc}
\toprule  
Method& \makecell[c]{$\errR{}$\\ (degrees)}& \makecell[c]{$\errt{}$ ($\cdot 10^{-1}$)\\ ($\ell_1$, $\ell_2$)} & \makecell{$\errpw{}$ ($\cdot 10^{-1}$)\\($\ell_1$, $\ell_2$)}\\
\midrule  
DCP-GT& 3.841 & 0.061, 0.039& 0.068, 0.046\\ 
DCP-CD&  7.021 &0.185, 0.114&0.193, 0.130\\ 
DCP-Ours& \textbf{4.841}& \textbf{0.07, 0.046}& \textbf{0.079}, \textbf{0.054}\\ 
\midrule  
FMR-GT& 2.122& {0.058, 0.039}& {0.064, 0.043} \\
FMR-CD&  6.207 &0.187, 0.123 &0.228, 0.134\\
FMR-Ours& \textbf{1.521}& \textbf{0.089, 0.051}& \textbf{0.091, 0.063}\\
\midrule 
RPM-GT& 1.921& 0.030, 0.021 &0.033, 0.023\\
RPM-CD& 8.373 &0.197, 0.16 &0.193,0.133\\
RPM-Ours& \textbf{1.33}& \textbf{0.032, 0.024} &\textbf{0.034, 0.023}\\
\bottomrule 
\end{tabular}}
\end{center}
\label{tab:Human dataset}
\end{table}

Different from supervised learning, training an unsupervised model often requires suitable initialization~\cite{feng2021recurrent}. We use the following steps to derive initialization for the synthetic datasets. During preprocessing, we first generate an easier dataset with 100 data pairs and smaller pose differences between the source and target point clouds, and train the model on them for 500 epochs to obtain an overfit model. Then we train the model on $10\%$ of the training dataset. Finally, we train the model on the whole training dataset with a reduced learning rate. For the real dataset, we generate an easier dataset consisting of pairs that are separated by a smaller number of frames during preprocessing, while the remaining training process is the same as the synthetic datasets.

\begin{figure}[t]
\begin{center}
       \includegraphics[width=\linewidth]{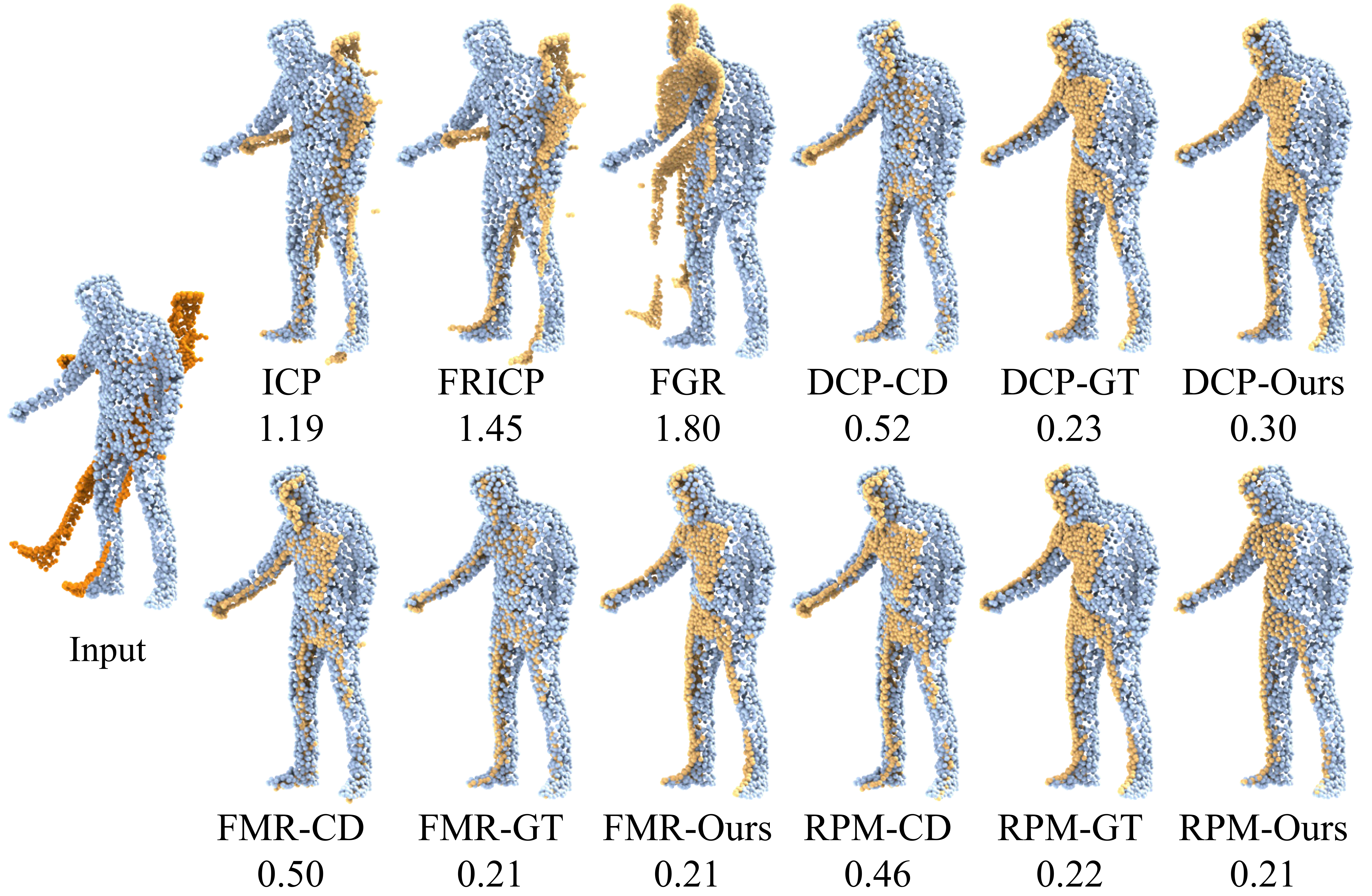}
\end{center}
    \caption{Examples of registration results using different methods on the Human dataset~\cite{Axyz}. 
}
    \label{fig:Human dataset results}
 \end{figure}
 
\begin{figure}[t]
\begin{center}
       \includegraphics[width=\linewidth]{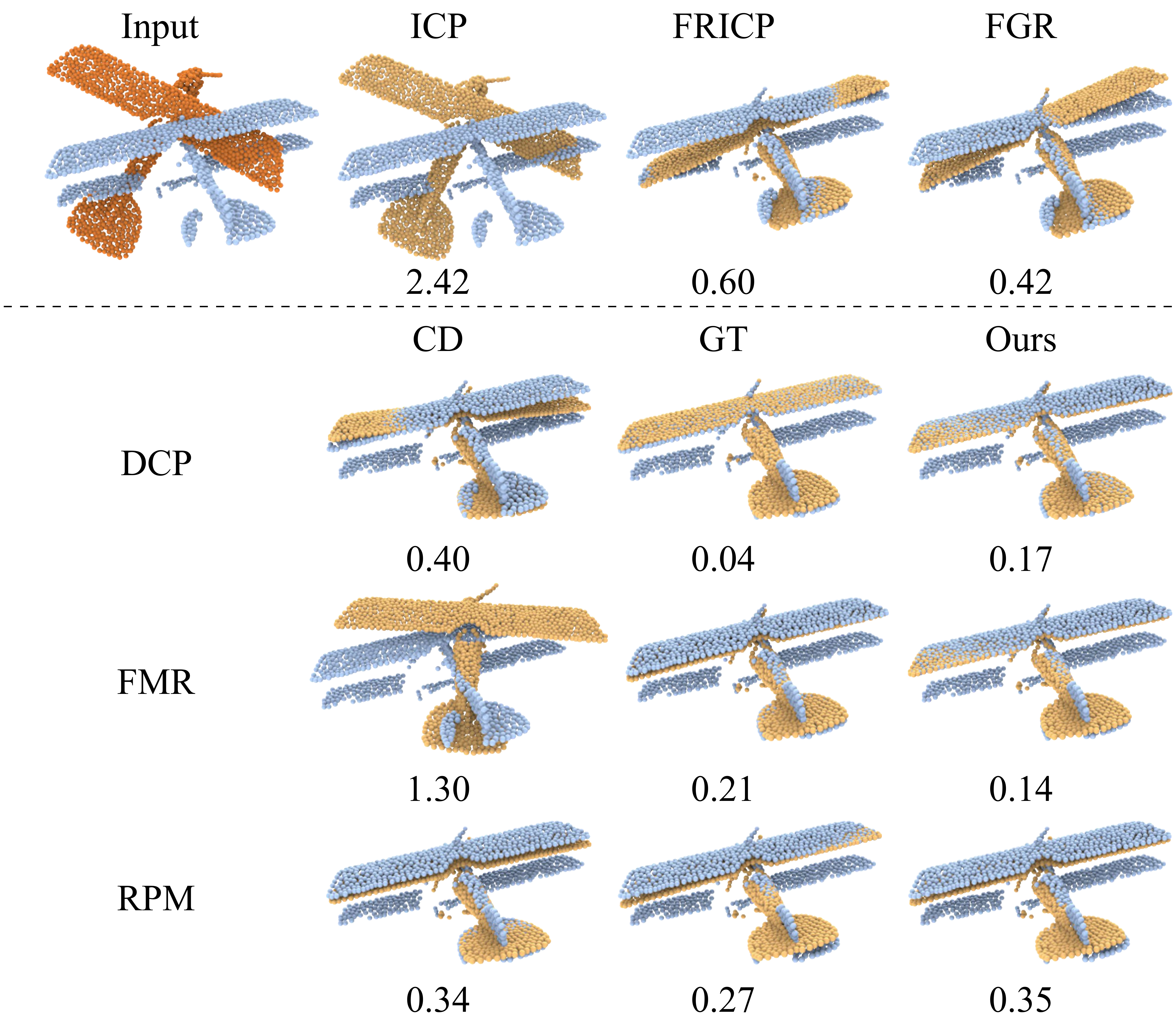}
\end{center}
    \caption{Examples of registration results using different methods on the Airplane dataset~\cite{Wu_2015_CVPR}.
    	 }
    \label{fig: Airplane dataset results}
 \end{figure}

\begin{table}[t]
	\caption{Comparison between different optimization-based and unsupervised learning methods on the real dataset.}
\begin{center}
    \label{tab:Real dataset}
\resizebox{0.46 \textwidth}{!}{\begin{tabular}{cccc}
\toprule  
Method& \makecell[c]{$\errR{}$\\ (degrees)}& \makecell[c]{$\errt{}$ ($\cdot 10^{-1}$)\\ ($\ell_1$, $\ell_2$)} & \makecell{$\errpw{}$ ($\cdot 10^{-1}$)\\($\ell_1$, $\ell_2$)}\\
\midrule  
ICP~\cite{Besl_ICP} & 17.98&0.337, 0.23& 0.238, 0.194 \\
FRICP~\cite{Zhang:2021:FRICP} & 11.08&0.199, 0.139& 0.151, 0.112\\
FGR~\cite{Zhou_2016_FGR} &12.79&0.211, 0.142&0.260, 0.21 \\
\midrule  
FMR-CD& 7.559  &0.469, 0.34& 0.531, 0.384\\
FMR-Ours& \textbf{3.263}& \textbf{0.089, 0.065}& \textbf{0.101}, \textbf{0.075}\\
\midrule 
RPM-CD& 11.28 & 0.342, 0.246&0.376, 0.272\\
RPM-Ours& \textbf{2.972}& \textbf{ 0.057, 0.04} &\textbf{0.068, 0.05}\\
\bottomrule 
\end{tabular}}
\end{center}
\label{tab:real datasets}
\end{table}

\begin{figure}[t]
	\begin{center}
		\includegraphics[width=\linewidth]{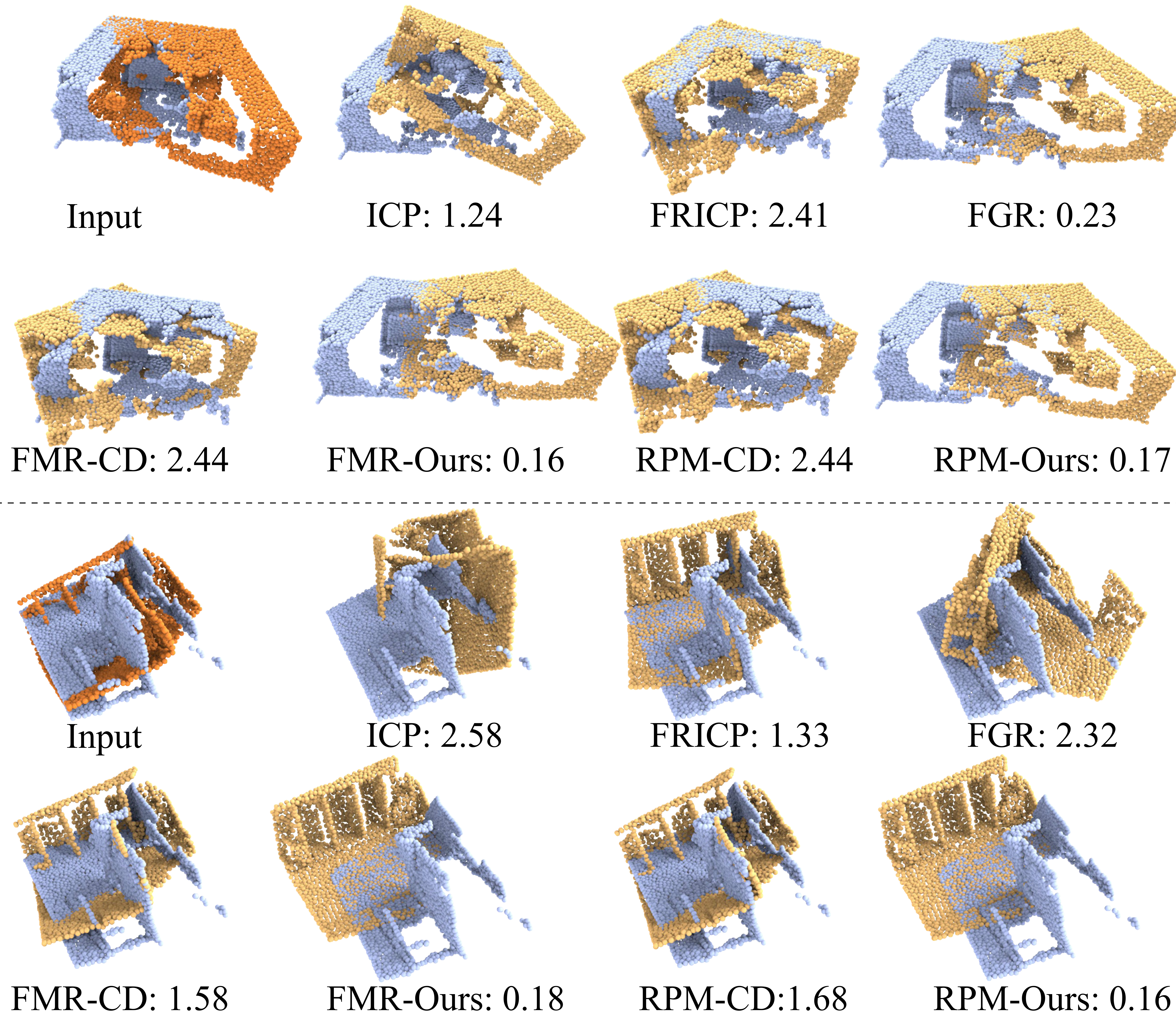}
	\end{center}
	\caption{Comparison of different methods on the Real dataset with deep learning frameworks with different metrics.
	}
	\label{fig: real dataset results}
\end{figure}

Tab.~\ref{tab:Airplane dataset} and Tab.~\ref{tab:Human dataset} show the performance results on the Airplane dataset and the Human dataset, respectively. They show that our metric is suitable for unsupervised deep learning frameworks, with superior performance compared to unsupervised variants with Chamfer distance, and even better performance than the supervised versions in some cases. Especially for FMR, using our metric to their original unsupervised framework greatly improves the performance. Figs.~\ref{fig:Human dataset results} and \ref{fig: Airplane dataset results} show registration results using different methods on two problems from the Human dataset and the Airplane dataset respectively. For both problems, optimization-based methods converge to local minima. Unsupervised learning approaches using our metric produce much better alignment than their counterparts using Chamfer distance.

Tab.~\ref{tab:real datasets} and Fig.~\ref{fig: real dataset results} compare the performance of different methods on our real dataset. Due to the lack of ground-truth alignment labels, supervised learning approaches are no longer applicable. 
As Fig.~\ref{fig: real dataset results} shows that ICP-based methods~\cite{Zhang:2021:FRICP,Besl_ICP} are prone to local minima due to sensitivity to initial values, whereas the feature-based method of~\cite{Zhou_2016_FGR} performs poorly due to the noisy normals in real data. Unsupervised learning using our metric is more robust than the Chamfer distance and performs the best in this benchmark.

\begin{figure}[t]
	\begin{center}
	\includegraphics[width=1.0\columnwidth]{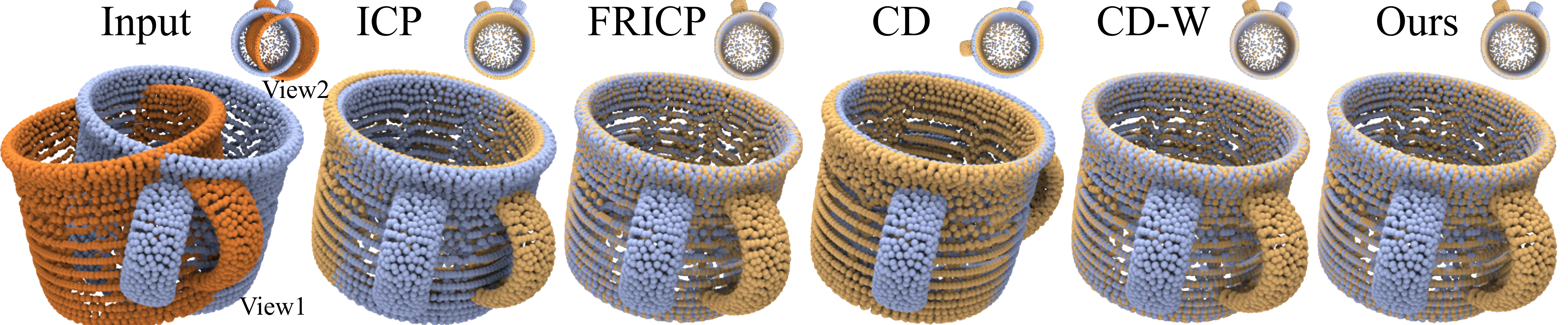}
	\end{center}
	\caption{A failure case of optimization using our metric, as well as results from other methods.}
	\label{fig:failure_case}
\end{figure}

\section{Conclusion and Future Work}
We have proposed a novel metric for point cloud registration. The main contributions of our work are two aspects. First, the proposed metric is based on intersections of uniformly random straight lines set in space, which can obtain richer information and more likely to achieve the global minimum. Second, our proposed metric can turn various supervised learning frameworks into unsupervised and has the ability to train on massive real unlabeled suitable data sets. Extensive ablation studies have verified the effectiveness of each component of our metric. Experiments on synthetic and real datasets show that our metric is competitive compared to the previous metrics and can be used in the loss function of deep learning frameworks.

Fig.~\ref{fig:failure_case} shows a failure case for our metric: after the symmetric bodies of two point clouds of a mug are aligned, our method is unable to align the handles. This is because within a set of random sample straight lines, only a small number of them will hit the handles, and their alignment effect will be dominated by other lines that tend to maintain the current alignment between the mug bodies. 

In the future, a possible avenue of research is to further investigate why the proposed metric can achieve better performance. Our conjecture is that the intersection with random straight lines introduce randomness to the optimization process and help the solver escape from local minima. This has been observed in our experiments, but will need more rigorous investigation to verify and understand. Another potential direction is to use relevant mathematical theories such as integral geometry to interpret our metric, which can be a challenging and interesting future work. 

{\small{\paragraph{Acknowledgement} This work was supported by NSFC (No.~62122071), the Youth Innovation Promotion Association CAS (No.~2018495), ``the Fundamental Research Funds for the Central Universities'' (No.~WK3470000021), and Guangdong International Science and Technology Cooperation Project (No.~2021A0505030009).}}

{\small
\bibliographystyle{ieee_fullname}
\bibliography{egbib}
}
\clearpage
\begin{strip}
   \begin{center}
       \textbf{\Large Supplementary Material}
   \end{center}
\end{strip}

This supplementary material shows some details that were not given in the paper due to constraints of space, including adding an experiment about the variants of Chamfer distance based on Welsch's function~\cite{Paul_Robust_Welsch}, more visualization of our comparison.

\subsection*{Comparison with Welsch's Chamfer Distance}
As a supplement, we also add an experiment, which compares with variants of the chamfer distance with robust Welsch's function.
\begin{equation*}
\label{Eq:welsch chamfer}
f(\widetilde{\mathcal{S}}_{(\mathbf{R}, \mathbf{t})}, \mathcal{T})_{welsch}=\sum_{(\mathbf{x},\mathbf{y})\in C}D_{wel}(\mathbf{Rx+t},\mathbf{y}),
\end{equation*}
\begin{equation*}
D_{wel}(\mathbf{x}, \mathbf{y}) =  \psi_{\nu}(\|\mathbf{x}-\mathbf{y}\|_{2}), 
\label{Eq:sparse_metric_chamfer}
\end{equation*}
where $C$ is a set of closest corresponding pairs, and  $\mathbf{x}$ and $\mathbf{y}$ are the sample points on $\widetilde{\mathcal{S}}$ and $\mathcal{T}$ respectively, and $D(\cdot, \cdot)$ is a distance metric between each corresponding point pair, $\psi_{\nu}(x)= 1-\exp(-\frac{x^2}{2{\nu}^{2}})$ is Welsch's function to reduce the influence of corresponding pairs with long distances. Here $\nu$ is a hyperparameter that determine the sparseness. We set $\nu =\nu_{0}d_{\text{med}}$, where ${d}_{\text{med}}$ represents the median value of all corresponding pairs' distances and $\nu_{0}$ is a hyperparameter.

\begin{figure}[t]
    \begin{center}
       \includegraphics[width=\linewidth]{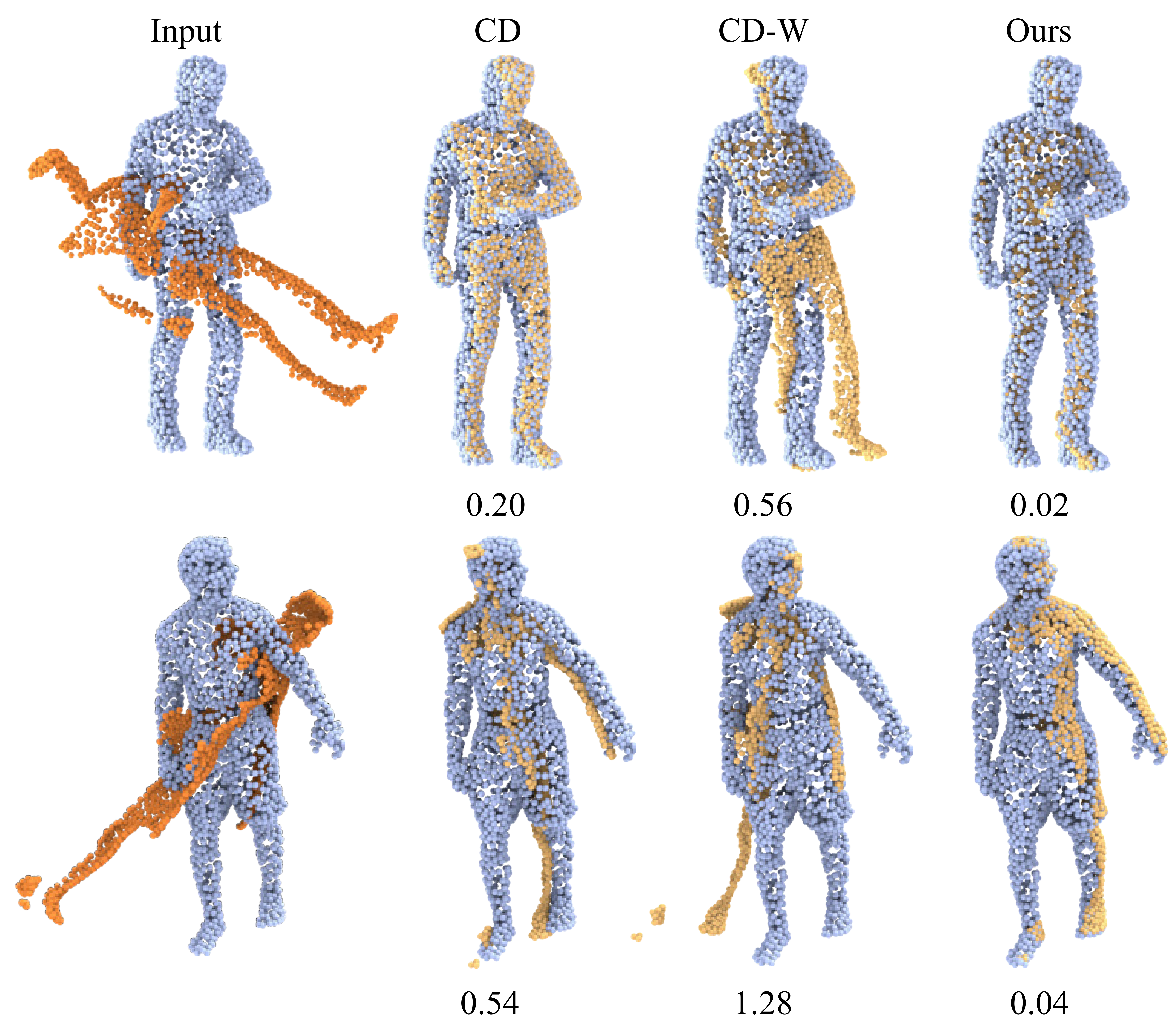}
     \end{center}
    \caption{Comparison of different variants of chamfer distance on the Human dataset.
    }
    \label{fig: support chamfer human}
 \end{figure}

\begin{figure}[t]
    \begin{center}
       \includegraphics[width=\linewidth]{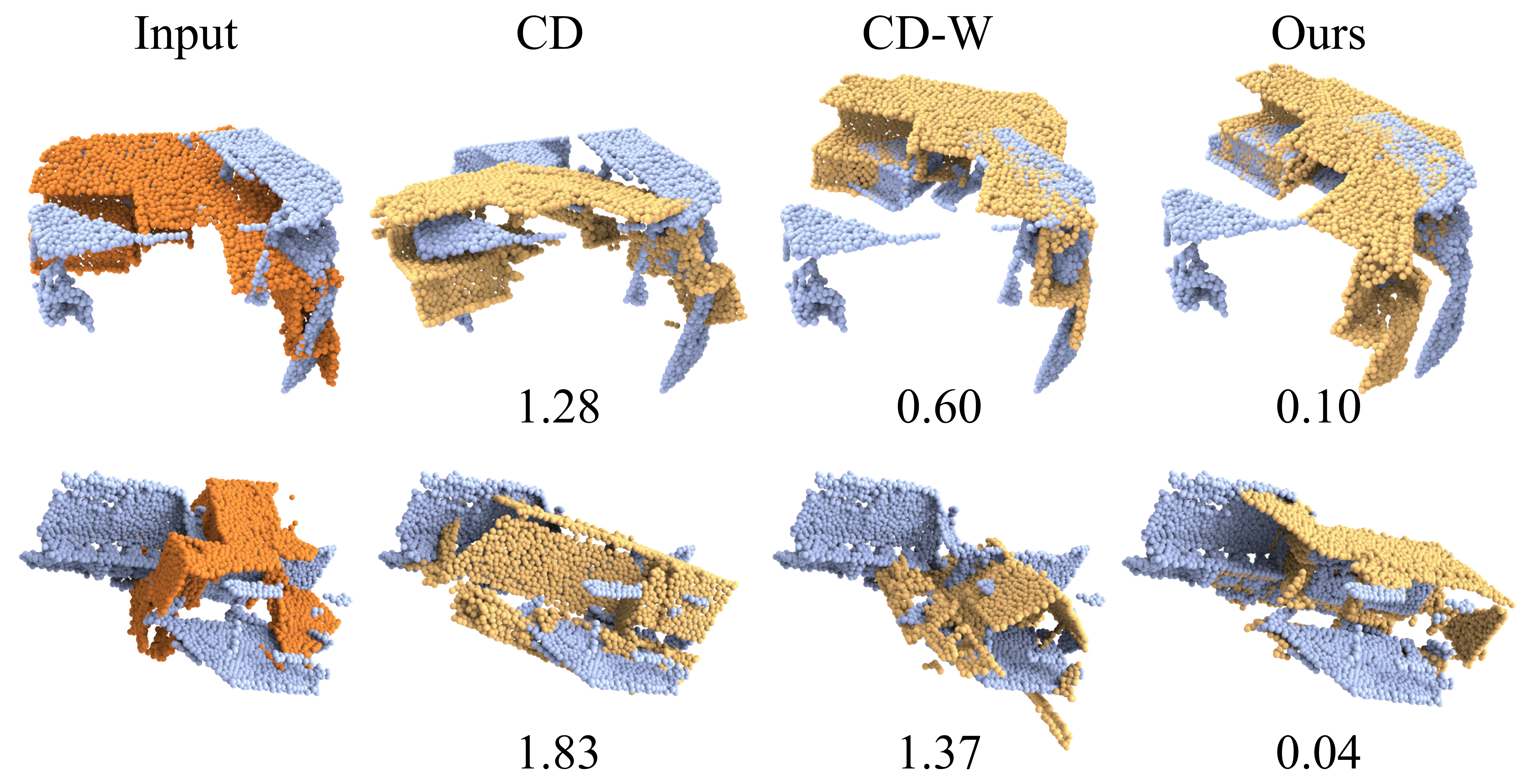}
     \end{center}
    \caption{Comparison of different variants of chamfer distance on 3D-Macth datasets.
    }
    \label{fig: support chamfer real }
 \end{figure}

 We optimize the Lie algebraic representation of rigid transformation with our metric and variants of chamfer distance by gradient descent method~\cite{NoceWrig06}, and use the Human test dataset as the benchmark in this experiment; we consider the influence of hyperparameters, and thus choose three sets of parameters, which are the $\nu_{0}=2,\nu_{0}=0.5,\nu_{0}=0.1$. The Tab.~\ref{tab:Comparison with welsch chamfer} indicates that our metric is superior to variants of chamfer distance based on Welsch's function, and the Fig.~\ref{fig: support chamfer human} and Fig.~\ref{fig: support chamfer real} are the visualized results of variants of Chamfer distance, and our metric on the Human dataset~\cite{Axyz} and the 3d-Match dataset~\cite{zeng20163dmatch}, respectively. It shows our metric can generate more robust and global optimal results.

\begin{table}[h]
\begin{center}
	\caption{Comparison with variants of chamfer distance by directly optimizing a Lie algebra  on Axyz-pose human dataset \cite{Axyz}. It shows the rotation errors(degree), translation errors and piecewice errors defined in Eq.~\eqref{Eq:eval_metric}.}
\resizebox{0.46 \textwidth}{!}{\begin{tabular}{ccccc}
\toprule  
Method& \makecell[c]{$\errR{}$\\ (degrees)}& \makecell[c]{$\errt{}$ ($\cdot 10^{-1}$)\\ ($\ell_1$, $\ell_2$)} & \makecell{$\errpw{}$ ($\cdot 10^{-1}$)\\($\ell_1$, $\ell_2$)}\\
\midrule  
CD&  5.863 &0.148, 0.132&0.151, 0.14\\
\midrule  
CD-W ($\nu_{0} = 0.1$)& {12.574}& {0.196, 0.152}& {0.384, 0.304}\\
CD-W ($\nu_{0} = 0.5$)& {4.84}& { 0.086, 0.078}&{0.108, 0.087}\\
CD-W ($\nu_{0} = 2$)& {1.4}& {0.0267, 0.024}& {0.051, 0.043}\\
\midrule
Ours ($\nu_{0}=0.5$)& \textbf{0.576}& \textbf{0.017, 0.013}& \textbf{0.018, 0.015}\\
\bottomrule 
\end{tabular}}
\end{center}
 \label{tab:Comparison with welsch chamfer}
\end{table}
\subsection*{More Visualized Results Compared with Traditional Methods}
\begin{figure}[t]
    \begin{center}
       \includegraphics[width=\linewidth]{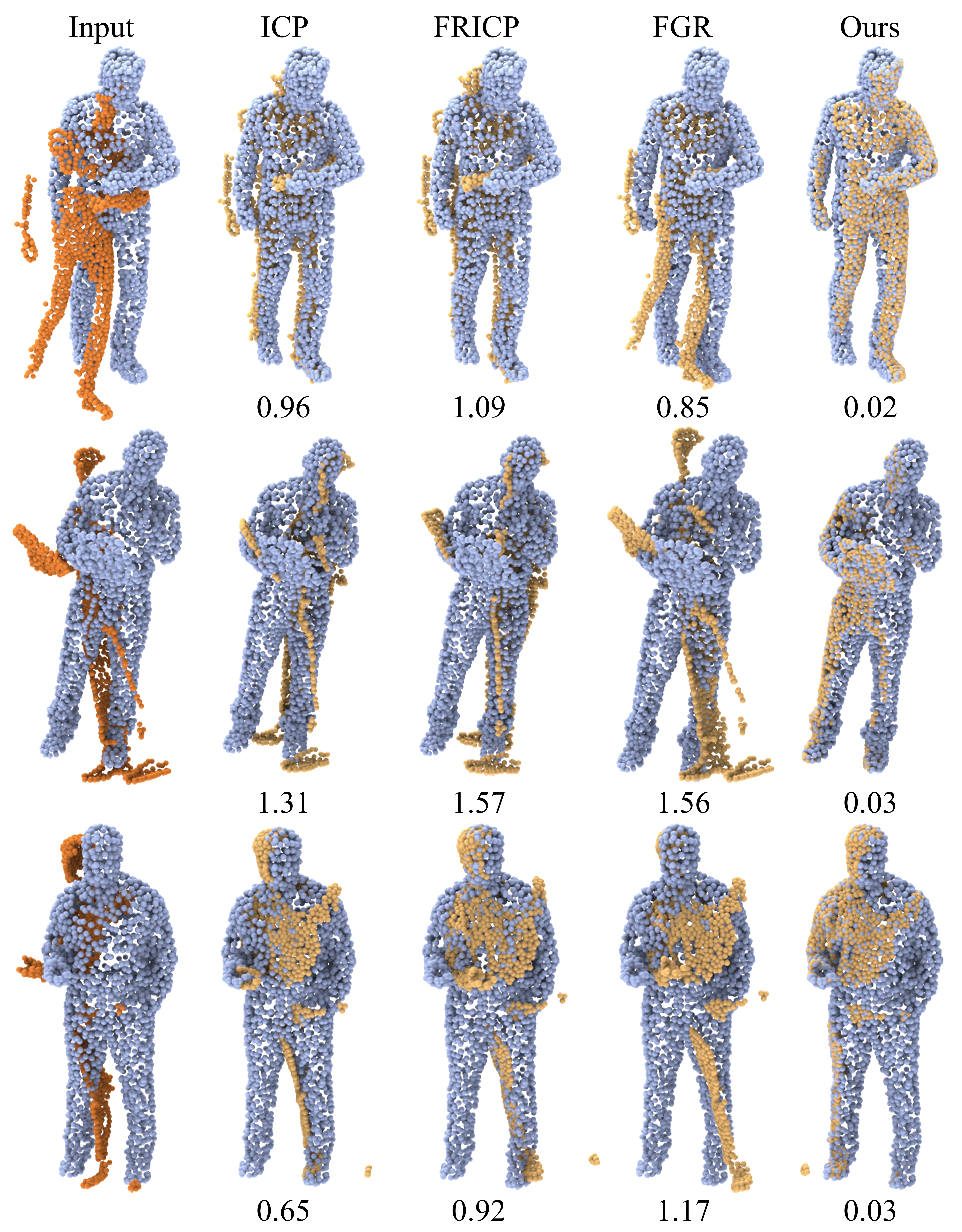}
     \end{center}
    \caption{Comparison with different traditional methods on the Human dataset.
    }
    \label{fig: support comparison with traditional method human }
 \end{figure}

\begin{figure}[t]
    \begin{center}
       \includegraphics[width=\linewidth]{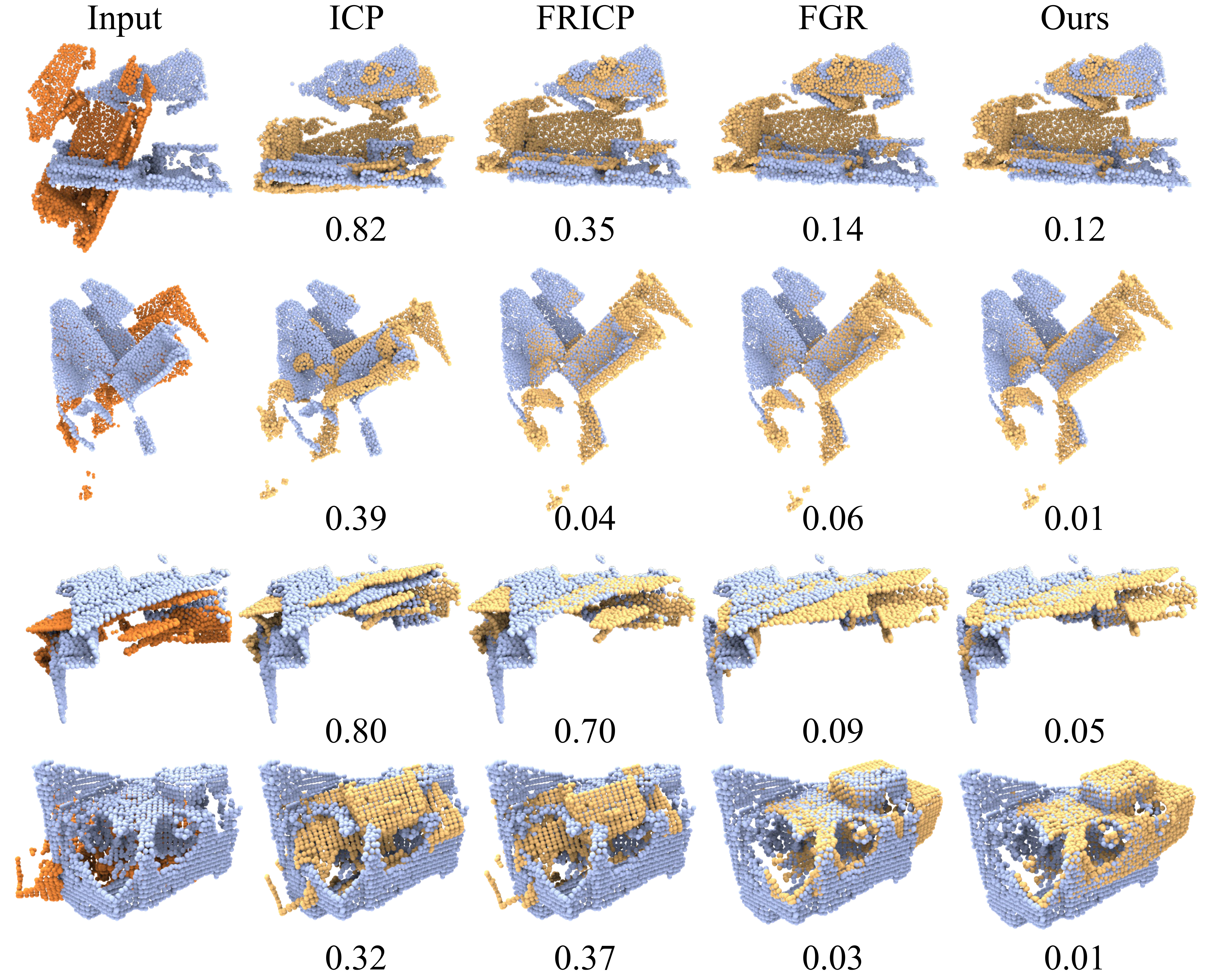}
     \end{center}
    \caption{Comparison with different traditional methods on the 3d-Match dataset. 
    }
    \label{fig: support comparison with traditional method real }
 \end{figure}

We provide more visualized results compared with traditional methods, the Fig.~\ref{fig: support comparison with traditional method human } and Fig.~\ref{fig: support comparison with traditional method real } are the visualized results of other traditional methods and our metric on the Human dataset and the 3D-Match dataset, respectively. The Fig.~\ref{fig: support comparison with traditional method human } shows our metric is easier to jump out of the local optimal solution to attain the global optimal solution, but ICP-based methods has reached the local optimum. The Fig.~\ref{fig: support comparison with traditional method real } indicates that our metric can also achieve better results for the real data.

\begin{table}[!t]
				\caption{The comparison of computation time(ms) per iteration(col 2-6) and of the run in an example(col 7-8) in FRICP~\cite{Zhang:2021:FRICP} and our method. In our method, we count the optimized time of two sub-processes per iteration: random straight line generation(Sam), finding the points of intersection, and expectation calculation(Inter). For the inference time of our method, we use DCP~\cite{Wang_2019_ICCV} to test. Since this time is related to the number of point clouds and random lines, we show the time under three different numbers of point clouds and two different numbers of random straight lines.}
	\begin{center}
		\resizebox{0.46 \textwidth}{!}{\begin{tabular}{c |c  c |c  c |c|| c|c}
				\multicolumn{1}{ c |}{\multirow{2}{*}{$\#$Points}}
				&\multicolumn{2}{ c  |}{$\#$Lines: 5000}
				&\multicolumn{2}{ c | }{$\#$Lines: 20000}
				&\multicolumn{1}{c||}{\multirow{2}{*}{{FRICP(S)}}}
				&\multicolumn{1}{c |}{\multirow{2}{*}{Inference}}
				&\multicolumn{1}{ c }{\multirow{2}{*}{{FRICP(W)}}}
				\\\cline{2-5}
				\multicolumn{1}{ c| }{}
				&Sam  	&Inter	
				&Sam  	&Inter 
				&\multicolumn{1}{ c||}{}
				&\multicolumn{1}{ c|}{}\\\hline
				1024& {36.6}& {50.4}& {46.4}& {204.4}&{0.3}&{19.7}&{43.9}\\
				\hline
				5000& {42.3}& {145.5}& {45.6}& {236.2}&{1.9}&{20.7}&{223.3}\\
				\hline
				10000& {45.6}& {191.9}& {47.4}& {544.4}&{3.2}&{22.4}&{379}\\
			\end{tabular}}
			\label{tab:Comparison Cost}
		\end{center}
	\end{table}

\section*{Computation Cost of Our Metric}
Tab.~\ref{tab:Comparison Cost} shows the training and inference time of our metric per iteration and compares them with the average computational time for FRICP, and model-based inference is very fast. Although our models need take a longer time to train, after the training, the actual registration only involves inference which is much more efficient. Besides, our current metric is implemented with Pytorch~\cite{NEURIPS2019_9015}. If in the GPU state, the general setting requires about 15G of memory, which is also an interesting direction that can be improved in the future.

\section*{More Details of Our Experiments Setting}
\paragraph{RPM-net framework}~\cite{yew2020-RPMNet}\footnote{\url{https://github.com/yewzijian/RPMNet}} used a less sensitive initialization and more robust deep learning-based approach for rigid point cloud registration. The metrics which used is $l_{1}$ distance between the source point cloud $X$ transformed using the groundtruth transformation ${\mathbf{R}_{gt},\mathbf{t}_{gt}}$ and the predicted transformation ${\mathbf{R}_{pred},\mathbf{t}_{pred}}$.
\begin{equation*}
\begin{aligned}
L_{reg}=\frac{1}{J}\sum_{j}^{J}|(\mathbf{R}_{gt}x_{j}+\mathbf{t}_{gt})-(\mathbf{R}_{pred}\mathbf{x}_{j}+\mathbf{t}_{pred})|.
\end{aligned}
\end{equation*}
\begin{equation*}
\begin{aligned}
L_{inlier}=-\frac{1}{J}\sum_{j}^{J}\sum_{k}^{K}m_{jk}-\frac{1}{K}\sum_{k}^{K}\sum_{j}^{J}m_{jk}.
\end{aligned}
\end{equation*}
The overrall loss is the weighted sum of the two losses:
$L_{total}=L_{reg}+\lambda L_{inlier}$
where we use $\lambda =0.01$ in all our experiments for supervised learning, and use our metric and Chamfer distance 
replace the $L_{reg}$ for our experiment setting and Chamfer distance experiment setting, respectively.
We compute the loss for every iteration $i$, but weigh the losses by ${\frac{1}{2}}^{(N_{i}-i)}$ to 
give later iterations higher weights, where $N_{i}$ is the total number of iteration during training.

\paragraph{DCP framework}~\cite{Wang_2019_ICCV}\footnote{\url{https://github.com/WangYueFt/dcp}} is a learning method based on differential SVD, here the loss function to measure the model's agreement to the ground-truth rigid motions:
\begin{equation*}
\begin{aligned}
Loss={||\mathbf{R}^{T}_{xy}\mathbf{R}^{g}_{xy}-I||} + {||\mathbf{t}_{xy}-\mathbf{t}_{xy}^{g}||^{2}} +\lambda{||\theta||}^{2}.
\end{aligned}
\end{equation*}

Here, $g$ denotes ground-truth. The first two terms define a simple distance on $SE(3)$. The third term denotes 
Tikhonov regularization of the DCP parameters $\theta$, which reduce the complexity of the network. We replace 
the first two terms with our metric and Chamfer distance as our experiment setting and Chamfer distance experiment setting, respectively.

\paragraph{FMR framework}~\cite{Huang_2020_CVPR}\footnote{\url{https://github.com/XiaoshuiHuang/fmr}} is a fast semi-supervised approach for robust point cloud registration without correspondence.
The loss functions in their paper are the 
\begin{equation*}
\begin{aligned} 
loss=loss_{cf}+loss_{pe}.
\end{aligned}
\end{equation*}

\begin{equation*}
\begin{aligned}
loss_{cf}=\sum_{p\in A}\sum_{i=1}^{N}\mathop{min}\limits_{q\in S^{*}}{||\phi_{\theta_{i}(p;x)}-q||}_{2}^{2}               \\
+\sum_{q\in S^{*}}\mathop{min}\limits_{i,\in 1\dots N}\mathop{min}\limits_{p \in A}{||\phi_{\theta_{i}}(p;x)-q||}^{2}_{2},
\end{aligned}
\end{equation*}
where $p \in A$ is a set of points sampled from a unit square $[0, 1]^2$, $x$ is a point cloud feature, $\phi_{\theta_{i}}$
is the $i^{th}$ component in the MLP parameters, $S^{\*}$ is the original input 3D points
\begin{equation*}
loss_{pe} = \frac{1}{M}\sum_{i=1}^{M}{||f(g_{est}\cdot{ }P)-f(g_{gt}\cdot{ }P)||}_{2}^{2},
\end{equation*}
where $P$ is a point cloud, and $M$ is its total point number.
For the unsupervised training, we only use the $loss_{cf}$; we replace the $loss$ with our metric as our experiment setting.
And we significantly improve the performance of the unsupervised manners.

\section*{Extend Our Metric to SVD Solver}
Regarding the discrete version of Eq.~\eqref{Eq:our_metric}:
\begin{equation*}
\label{Eq:our_metric}
f(\widetilde{\mathbf{X}}, \mathbf{Y})=\sum_{l\in A}w_l(\sum_{\mathbf{x}_i^l\in\mathbf{X}_l}
{D(\widetilde{\mathbf{x}}_i^l, \mathbf{y}_{{\pi_i}^{l}}^l)}+\sum_{\mathbf{y}_j^l\in\mathbf{Y}_l} {D(\widetilde{\mathbf{x}}_{{\rho_j}^{l}}^l,\mathbf{y}_j^l)}).
\end{equation*}
As described in the FRICP~\cite{Zhang:2021:FRICP}, based on the corresponding points  $\{(\mathbf{x, y}), (\mathbf{x, y})\in C\}=\mathop{\cup}\limits_{l\in A}(\mathop{\cup}\limits_{x_{i}^{l} \in \mathbf{X}_{l}}(x_{i}^{l}, y_{{\pi}_{i}^{l}}^{l})\mathop{\cup}\limits_{y_{j}^{l} \in \mathbf{Y}_{l}}(x_{{\rho}_{j}^{l}}^{l}, y_{j}^{l}))$ between the source intersections $\mathop{\cup}\limits_{l\in A} \mathbf{X}_{l}$ and target intersections $\mathop{\cup}\limits_{l\in A} \mathbf{Y}_{l}$.
We can extend our non-linear metric into quadratic by replacing the Welsch's function with quadratic surrogate function, then, we get the sum of squared distance between the points $\mathbf{x, y}$.
\[
(\mathbf{R}^{*},\mathbf{t}^{*})=\mathop{\arg\min}\limits_{(\mathbf{R},\mathbf{t})}\sum_{\mathbf{(x, y)}\in\mathbf{C}}{w_{\mathbf{(x, y)}}||(\mathbf{R}{\mathbf{x}}+\mathbf{t}- \mathbf{y})||}_{2}^{2},
\]
where $w_{\mathbf{(x, y)}}= \psi_{\nu}(\|\mathbf{x}-\mathbf{y}\|_{2})*w_{l}$. It can be solved in closed form via SVD~\cite{sorkine2017least}, which implemented with Pytorch~\cite{NEURIPS2019_9015}.
\end{document}